\documentclass[runningheads]{llncs}

\usepackage[T1]{fontenc}

\usepackage{graphicx}
\usepackage{comment}
\usepackage{amsmath,amssymb}
\usepackage{color}
\usepackage{url}
\newcommand{\mypara}[1]{\noindent{\bf{#1}}}
\usepackage{amsmath,amsfonts}
\usepackage{algorithmic}
\usepackage{algorithm}
\usepackage{array}
\usepackage{textcomp}
\usepackage{stfloats}
\usepackage{url}
\usepackage{verbatim}
\usepackage{graphicx}
\usepackage{cite}
\usepackage{xspace}
\usepackage{bm}
\usepackage{wrapfig}
\usepackage{xurl}

\usepackage{multirow}
\usepackage{booktabs}

\newcolumntype{H}{>{\setbox0=\hbox\bgroup}c<{\egroup}@{}}
\usepackage[pagebackref=true,breaklinks=true,colorlinks,bookmarks=false]{hyperref}
\usepackage{cleveref}
\usepackage{orcidlink}

\usepackage{xcolor}
\newcommand{\ba}{{\textit{B}}\xspace}
\newcommand{\nl}{{\textit{N}}\xspace}

\newcommand{\ourmodel}{\textsc{AV-Diff}\xspace}
\newcommand{\TCAF}{\textsc{TCaF}\xspace}
\newcommand{\activity}{{ActivityNet-FSL}\xspace}
\newcommand{\ucf}{{UCF-FSL}\xspace}
\newcommand{\vgg}{{VGGSound-FSL}\xspace}
\renewcommand\UrlFont{\color{blue}\rmfamily}

\newif\ifreview
\reviewfalse

\ifreview
	\usepackage{lineno}
	\renewcommand\thelinenumber{\color[rgb]{0.2,0.5,0.8}\normalfont\sffamily\scriptsize\arabic{linenumber}\color[rgb]{0,0,0}}
	\renewcommand\makeLineNumber {\hss\thelinenumber\ \hspace{6mm} \rlap{\hskip\textwidth\ \hspace{6.5mm}\thelinenumber}} 
	\linenumbers
\fi

\begin{document}


\def\SubNumber{102}

\def\GCPRTrack{Main Track}

\title{Text-to-feature diffusion for audio-visual few-shot learning}

\ifreview
	\titlerunning{GCPR 2023 Submission \SubNumber{}. CONFIDENTIAL REVIEW COPY.}
	\authorrunning{GCPR 2023 Submission \SubNumber{}. CONFIDENTIAL REVIEW COPY.}
	\author{GCPR 2023 - \GCPRTrack{}}
	\institute{Paper ID \SubNumber}
\else

	\author{Otniel-Bogdan Mercea\inst{1}\orcidlink{0000-0002-3586-1703}\index{Mercea, Otniel-Bogdan} \and
	Thomas Hummel\inst{1}\orcidlink{0000-0003-3201-360X} \and
	A. Sophia Koepke\inst{1}\orcidlink{0000-0002-5807-0576}\index{Koepke, A. Sophia} \and Zeynep Akata\inst{1,2}\orcidlink{0000-0002-1432-7747}}
	
	\authorrunning{O.-B. Mercea et al.}
	
	\institute{University of T{\"u}bingen \and MPI for Intelligent Systems \\
	\email{\{otniel-bogdan.mercea, thomas.hummel, a-sophia.koepke,zeynep.akata\}@uni-tuebingen.de}}
\fi

\maketitle              

\begin{abstract}
Training deep learning models for video classification from audio-visual data commonly requires vast amounts of labelled training data collected via a costly process. 
A challenging and underexplored, yet much cheaper, setup is few-shot learning from video data. In particular, the inherently multi-modal nature of video data with sound and visual information has not been leveraged extensively for the few-shot video classification task.
Therefore, we introduce a unified audio-visual few-shot video classification benchmark on three datasets, i.e.\ the \vgg, \ucf, \activity  datasets, where we adapt and compare ten methods. In addition, we propose \ourmodel, a text-to-feature diffusion framework, which first fuses the temporal and audio-visual features
via cross-modal attention and then generates multi-modal features for the novel classes. We show that \ourmodel obtains state-of-the-art performance on our proposed benchmark for audio-visual (generalised) few-shot learning. Our benchmark paves the way for effective audio-visual classification when only limited labelled data is available.
Code and data are available at \url{https://github.com/ExplainableML/AVDIFF-GFSL}.

\end{abstract}
\section{Introduction}
The use of audio-visual data can yield impressive results for video classification \cite{nagrani2021attention, xiao2020audiovisual, patrick2020multi}. The complementary knowledge contained in the two modalities results in a richer learning signal than using unimodal data. 
However, video classification frameworks commonly rely on significant amounts of costly training data and computational resources. 
To mitigate the need for large amounts of labelled data, we consider the few-shot learning (FSL) setting where a model is tasked to recognise new classes with only a few labelled examples. 
Moreover, the need for vast computational resources can be alleviated by operating on the feature level, using features extracted from pre-trained visual and sound classification networks.

In this work, we tackle the task of few-shot action recognition in videos from audio and visual data which is an understudied problem in computer vision. In the few-shot setting, a model has to learn a transferable audio-visual representation which can be adapted to new classes with few annotated data samples. In particular, we focus on the more practical generalised FSL (GFSL) setting, where the aim is to recognise samples from both the base classes, i.e.\ classes with many training samples, and from novel classes which contain only few examples. 
Additional modalities, such as text and audio, are especially useful for learning transferable and robust representations from few samples. 

\begin{wrapfigure}{r}{0.52\textwidth}
    \centering
    \includegraphics[width=0.52\textwidth]{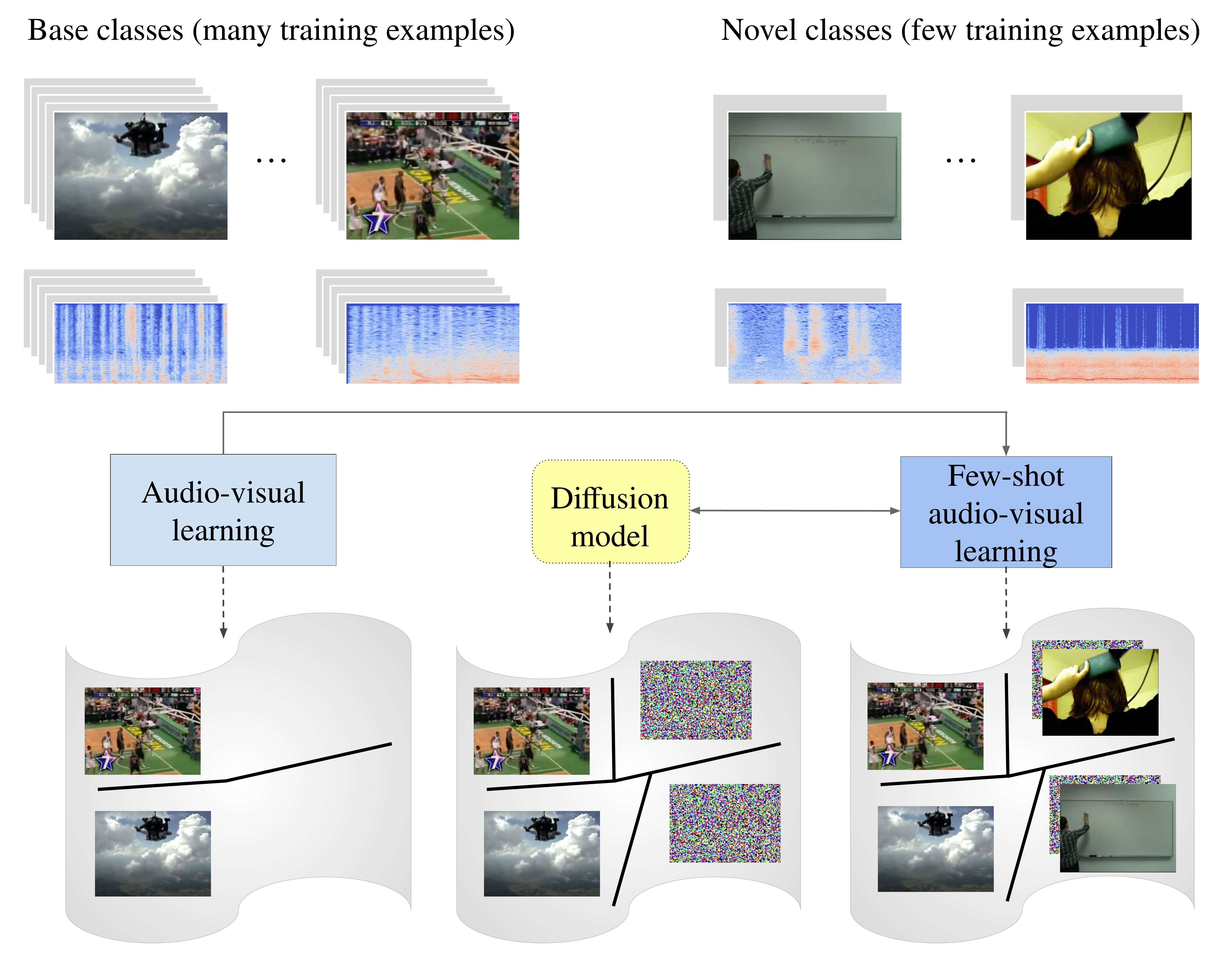}
    \caption{\ourmodel learns to fuse the audio-visual inputs into multi-modal representations in the audio-visual learning stage (left). In the few-shot learning stage (right), the multi-modal representations from the previous stage are used to concurrently train (double arrow line) a text-conditioned diffusion model on all the classes (middle) and a classifier. The classifier is trained on real features from base classes and real and synthetic features from novel classes. 
     } 
    \label{fig:teaser}
    
\end{wrapfigure}
To the best of our knowledge, the FSL setting with audio-visual data has only been considered for speech recognition~\cite{zhang2022audio}, and for learning an acoustic model of 3D scenes~\cite{majumder2022fewshot}. 
Moreover, existing video FSL benchmarks are not suitable for the audio-visual setting. In particular, the SomethingV2 and HMDB51 benchmarks proposed in \cite{cao2020few} and \cite{zhang2020few} do not contain audio and about 50\% of the classes in the UCF101 benchmark from~\cite{xian2021generalized} have no sound either. The Kinetics split in \cite{zhu2018compound} suffers from an overlap with the classes used to pre-train the feature extractors \cite{xian2021generalized}, and \cite{xiao2020audiovisual, nagrani2021attention} show that the audio modality in Kinetics is less class-relevant than the visual modality. Existing audio-visual zero-shot learning benchmarks \cite{mercea2022avca, mercea2022tcaf} cannot directly be used for few-shot learning due to their distinct training and testing protocols. Moreover, the baselines in both settings differ significantly as state-of-the-art few-shot learning methods usually necessitate knowledge of novel classes through classification objectives and generative models, a condition that is not possible in zero-shot learning.
Thus, we introduce a new benchmark for generalised audio-visual FSL for video classification that is comprised of three audio-visual datasets and ten methods carefully adapted to this challenging, yet practical task.

To tackle our new benchmark, we propose \ourmodel which uses a novel hybrid cross-modal attention for fusing audio-visual information. Different to various attention fusion techniques in the audio-visual domain~\cite{nagrani2021attention, mercea2022tcaf, mercea2022avca} which use a single attention type or different transformers for each modality, our model makes use of a novel combination of within-modality and cross-modal attention in a multi-modal transformer. This allows the effective fusion of information from both modalities and across the temporal dimension of the inputs. 
 Furthermore, we introduce a novel text-conditioned diffusion model for generating audio-visual features to augment the few samples in the novel classes. In the image and video domain, generative adversarial networks (GANs) have been used to generate uni-modal features for data augmentation in the FSL setting~\cite{xian2021generalized, kumar2019protogan, xian2019f,narayan2020latent,hariharan2017low}.
However, we are not aware of prior works that have used diffusion models for multi-modal (audio-visual) feature generation in FSL.
Both, cross-modal fusion and text-to-feature diffusion contribute to significant boosts in performance on our proposed benchmark.

To summarise, our contributions are:
1) We introduce the audio-visual generalised few-shot learning task for video classification and a benchmark on three audio-visual datasets. We additionally adapt and compare ten methods for this task.
2) We propose a hybrid attention mechanism to fuse multi-modal information and a diffusion model for multi-modal feature generation to augment the training dataset with additional novel-class samples.
3) We obtain state-of-the-art performance across all three datasets, outperforming the adapted multi-modal zero-shot learning and video FSL models.

\section{Related work}

We discuss prior works in learning from audio-visual data, FSL, 
and feature generation in low-shot learning.

\mypara{Audio-visual learning.} Multi-modal inputs, such as audio and visual data, provide significantly more information than unimodal data, resulting in improved overall performance for video classification and acoustic scene classification~\cite{owens2016ambient,owens2018learning,alwassel2019self,patrick2020multi,korbar2018cooperative,aytar2016soundnet}. Approaches, such as \cite{fayek2020large,chen2021distilling}, use class-label supervision between modalities without requiring temporal alignment between the input modalities.
Besides audio and video classification, other domains also benefit from multi-modal data, such as lip reading ~\cite{afouras2020asr,afouras2018deep}, audio synthesis based on visual information \cite{zhou2019vision,goldstein2018guitar,koepke2020sight,su2020multi,gan2020foley,narasimhan2021strumming,koepke2019visual}, and localisation and separation of sounds in videos ~\cite{owens2018audio,tian2018audio,arandjelovic2018objects,gao2019co,chen2021localizing,Afouras20b,afouras2021selfsupervised}.  Recently, transformer models have gained popularity in audio-visual learning, e.g.\ for classification \cite{boes2019audiovisual}, event localization 
\cite{lin2020audiovisual}, dense video captioning \cite{iashin2020better}, and text-based video retrieval \cite{gabeur2020multi, wang2021t2vlad}. As shown in these works, transformers can effectively process multi-modal input. Thus, our proposed framework fuses audio-visual information using a transformer-based mechanism.

\mypara{FSL}
has been explored in the image domain~\cite{roy2022diffalign,sung2018learning, vinyals2016matching, hariharan2017low,qi2018low, ravi2016optimization, chen2019closer, wang2019simpleshot, douze2018low, wang2018low, liu2018learning, li2019learning, ye2020few, snell2017prototypical} and in the video domain~\cite{cao2020few,zhu2018compound, kim2022better, bishay2019tarn, xian2021generalized}. The popular meta-learning paradigm in FSL~\cite{cao2020few, zhu2018compound, bishay2019tarn,sung2018learning, vinyals2016matching,ravi2016optimization, wang2019simpleshot,liu2018learning,li2019learning, ye2020few}  
has been criticised by recent works \cite{xian2021generalized,chen2019closer, wang2019simpleshot,kangdecoupling}.
In the video domain, commonly a query and support set is used and each query sample is compared to all the support samples~\cite{zhu2018compound, bishay2019tarn, perrett2021temporal, cao2020few}. The number of comparisons grows exponentially with the number of ways and shots. These methods become prohibitively expensive for GFSL, where models are evaluated on both the base and the novel classes. Hence, we focus on the non-meta-learning approach in this work. Some non-meta-learning approaches have addressed the more challenging and practical GFSL setting for videos~\cite{xian2021generalized, kumar2019protogan} using unimodal visual data. In contrast, we propose to use multi-modal data in our novel (G)FSL benchmark for audio-visual video classification which provides the possibility to test a model in both scenarios (FSL and GFSL).

\mypara{Feature generation.}
Due to the progress of generative models, such as GANs~\cite{goodfellow2020generative,adler2018banach,mirza2014conditional, gatys2015neural,isola2017image} and diffusion models \cite{rombach2022high, esser2021imagebart, blattmann2022semi}, different works have tried to adapt these systems to generate features as a data augmentation mechanism. GANs have been used in zero-shot learning (ZSL) and FSL \cite{xian2019f, narayan2020latent,xian2021generalized, kumar2019protogan} to increase the number and diversity of samples, especially for unseen or novel classes.
Diffusion models have also been applied to image generation in the feature space, such as \cite{rombach2022high, vahdat2021score}, but not in the ZSL or FSL setting. It is known that GANs are hard to optimize~\cite{saxena2021generative} while diffusion models appear to be more stable, leading to better results~\cite{dhariwal2021diffusion}. Therefore, our proposed framework uses a text-conditioned diffusion model to generate features for the novel classes in the FSL setting.

\section{Audio-visual (G)FSL benchmark}
We describe the audio-visual (G)FSL setting, present our proposed benchmark that we construct from audio-visual datasets, and explain the methods that we used to establish baselines for this task.

\subsection{Audio-visual (G)FSL setting}\label{sec:problemsetting}
We address the tasks of (G)FSL using audio-visual inputs. The aim of FSL is to recognise samples from classes that contain very few training samples, so-called \textit{novel classes}. In addition, the goal of GFSL is to recognise both \textit{base classes}, which contain a significant amount of samples, and 
novel classes.

Given an audio-visual dataset $\mathcal{V}$ with $M$ samples and $C$ classes, containing base and novel classes, we have 
$\mathcal{V} = \{\mathcal{X}_{\bm{a}[i]}, \mathcal{X}_{\bm{v}[i]}, y_{[i]}\}_{i=1}^{M}$, 
where $\mathcal{X}_{\bm{a}[i]}$ represents the audio input, 
$\mathcal{X}_{\bm{v}[i]}$ the video input and 
$y_{[i]} \in \mathbb{R}^{C}$ the ground-truth class label. 
Both the audio and the video inputs contain temporal information.
Two frozen, pre-trained networks are used to extract features from the inputs,  VGGish~\cite{hershey2017cnn} for the audio features ${a}_{[i]} = \{a_{1}, \dots, a_{t}, \dots, a_{F_a}\}_i$ and  C3D~\cite{tran2015learning} for video features ${v}_{[i]} = \{v_{1}, ..., v_{t}, ..., v_{F_v}\}_i$. We use these specific feature extractors to ensure that there is no leakage to the novel classes from classes seen when training the feature extractors (Sports1M~\cite{karpathy2014sports} for the visual and YouTube-8M~\cite{abu2016youtube} for the audio modality), similar to \cite{mercea2022avca}. A potential leakage is harmful as it would artificially increase the performance and will not reflect the true performance.

All models are evaluated in the FSL and GFSL settings for $k$ samples in the novel classes (called shots), with $k \in \{1,5,10,20\}$.
During inference, in the FSL setting, the class search space is composed only of the novel class labels and the samples belonging to these classes.
In the GFSL setting, the search space contains both the novel and base class labels and their corresponding samples. 

Meta-learning approaches commonly use the notion of episodes, where each episode only uses $P$ novel classes randomly sampled from the total number of novel classes in a dataset, usually $P \in \{1,5\}$  (coined $P$-way).
However, similar to \cite{xian2021generalized}, we suggest using higher values for $P$ (e.g.\ all the classes in the dataset), so that the evaluation is closer to the real-world setting, as argued in \cite{xian2021generalized, hariharan2017low}. 
In our proposed FSL setting, $P$ corresponds to the total number of novel classes $P=N$, while for GFSL $P=C$. Our evaluation protocol is in line with \cite{hariharan2017low}.

\subsection{Dataset splits and training protocol}\label{sec:benchmarks}

We provide training and evaluation protocols for audio-visual (G)FSL along with splits for \ucf, \activity and \vgg. These are based on the UCF-101~\cite{soomro2012ucf101}, ActivityNet~\cite{caba2015activitynet} and VGGSound~\cite{chen2020vggsound} datasets. 

Our proposed training and evaluation protocol is similar to \cite{hariharan2017low, mercea2022avca, mercea2022tcaf}. The training protocol is composed of two stages, indicated by subscripts $_1$,$_2$. In the first stage, a model is trained on the training set $\textit{Train}_{1} = \mathcal{V}_{B_1} \cup \mathcal{V}_{N_1}$ where $\mathcal{V}_{B_1}$ consists of dataset samples from base classes, and $\mathcal{V}_{N_1}$ contains $k$ samples for each of the classes ${N_1}$. The trained model is then evaluated on $\textit{Val} = \textit{Val}_{B} \cup \textit{Val}_{N} $, where $\textit{Val}$ is the validation dataset which contains the same classes as $\textit{Train}_{1}$. 
In the first stage, the hyperparameters for the network are determined, such as the number of training epochs and the learning rate scheduler parameters. 

In the second stage, the model is retrained on the training set $\textit{Train}_{2}$, using the hyperparameters determined in the first stage. Here, $\textit{Train}_{2} = \mathcal{V}_{B_2} \cup \mathcal{V}_{N_2}$ with $\mathcal{V}_{B_2} = \textit{Train}_{1} \cup \textit{Val}$, and $\mathcal{V}_{N_2}$ contains $k$ samples for the novel classes in the $\textit{Test}$ set. The final model is evaluated on $\textit{Test}=\textit{Test}_B \cup \textit{Test}_N$ with $\textit{Train}_2 \cap \textit{Test} = \emptyset$. 
With a small number of shots, e.g.\ $k=1$, models risk a bias towards the novel samples in $\textit{Train}_2$. To obtain robust evaluation results, the second stage is repeated three times with $k$ randomly selected, but fixed samples from $\mathcal{V}_{N_2}$.
We provide dataset statistics in Table \ref{tab:datasets_statistics_table}.

\begin{table}[t]
    \centering
    \setlength{\tabcolsep}{2pt}
    \renewcommand{\arraystretch}{1.2}
    \caption{Statistics for our \vgg \textbf{(1)}, \ucf \textbf{(2)}, and \activity \textbf{(3)} benchmark datasets, showing the number of classes and videos in our proposed splits in the 5-shot setting. $\mathcal{V}_{B_1} \cup \mathcal{V}_{N_1}$ are used for training, $\textit{Val}_B$ and $\textit{Val}_N$ for validation in the first training stage. $\mathcal{V}_{B_2} \cup \mathcal{V}_{N_2}$ serves as the training set in the second stage, and evaluation is done on $\textit{Test}_B$ and $\textit{Test}_N$.  
    } 
    \resizebox{0.8\linewidth}{!}{%
    
    \begin{tabular}{l|c c c c | c c c c |c c c c}
    \toprule 
    & \multicolumn{4}{c|}{\textbf{$\#$ classes}} & \multicolumn{4}{c|}{\textbf{$\#$ videos} \textit{stage 1}} & \multicolumn{4}{c}{\textbf{$\#$ videos} \textit{stage 2}} \\
    & all & $\mathcal{V}_{B_1}$ & $\mathcal{V}_{N_1}$ & $\mathcal{V}_{N_2}$ & $\mathcal{V}_{B_1}$ & $\mathcal{V}_{N_1}$& $\textit{Val}_{B}$ & $\textit{Val}_{N}$ &$\mathcal{V}_{B_2}$ & $\mathcal{V}_{N_2}$ & $\textit{Test}_B$ & $\textit{Test}_N$  \\ \midrule
      
      \textbf{(1)} & 271 & 138 & 69 & 64 & 70351 & 345 &7817 & 2757 & 81270 & 320 & 9032 &2880\\
      
      \textbf{(2)} & 48 & 30 & 12 & 6 & 3174 & 60& 353 &1407 & 4994 & 30 & 555 &815\\
        
      \textbf{(3)} & 198 & 99 & 51 & 48 & 9204 &255 & 1023 &4052 & 14534 & 240& 1615 &  3812\\
      \bottomrule
     \end{tabular}
    }

    \label{tab:datasets_statistics_table}
   
\end{table}

\subsection{Benchmark comparisons}

To establish benchmark performances for the audio-visual GFSL task, we adapt ten recent state-of-the-art methods for video FSL from visual information only, from audio-visual representation learning, and from audio-visual ZSL.

We provide results with several few-shot video recognition frameworks that are adapted to the multimodal audio-visual setting.

\textbf{ProtoGan} \cite{kumar2019protogan} 
uses GANs conditioned on the visual prototypes of classes that are obtained by averaging the features of all videos in that class. We adapt it to audio-visual inputs by concatenating the visual and audio features before passing them into the model.

\textbf{SLDG} \cite{bo2020few} is a multi-modal video FSL that uses video frames and optical flow as input. It weighs the frame features according to normal distributions. We replace the optical flow in \cite{bo2020few} with audio features. 

\textbf{TSL} \cite{xian2021generalized} is the current state-of-the-art video FSL which uses a GAN to generate synthetic samples for novel classes. It does not fully use temporal information, as the final score is the average of scores obtained on multiple short segments. We adapt it to the multi-modal setting by concatenating input features from the audio and visual modalities.

Moreover, we have adapted audio-visual representation learning methods to the few-shot task as can be seen below.

\textbf{Perceiver}\cite{jaegle2021perceiver}, \textbf{Hierarchical Perceiver (HiP)}~\cite{carreira2022hierarchical}, and \textbf{Attention Fusion} \cite{fayek2020large} are versatile video classification methods and we provide comparisons with them. We use the implementations of the adapted Perceiver and Attention Fusion frameworks provided by \cite{mercea2022tcaf} and we implement HiP in a similar way. 

\textbf{MBT} \cite{nagrani2021attention} learns audio-visual representations for video recognition. It uses a transformer for each modality and these transformers can only exchange information using bottleneck attention. 

\textbf{Zorro}\cite{Recasens2023ZorroTM}, in contrast to MBT, uses two transformers that do not have access to the bottleneck attention. We adapt it by using a classifier on top of the averaged bottleneck attention tokens. 

Finally, we have adapted the state-of-the-art methods in the audio-visual zero-shot learning domain, as shown below.

\textbf{AVCA} \cite{mercea2022avca} is an audio-visual ZSL method which uses temporally averaged features for the audio and visual modalities. We adapt it by using a classifier on the video output, which is the strongest of the two outputs in \cite{mercea2022avca}.

\textbf{\TCAF} \cite{mercea2022tcaf} is the state-of-the-art audio-visual ZSL method. 
It utilizes a transformer architecture with only cross-modal attention, leveraging temporal information in both modalities. As it does not use a classifier, \TCAF outputs embeddings, and we determine the class by computing the distance to the semantic descriptors and selecting the closest one.

\section{\ourmodel framework}
\begin{figure*}[t]
    \centering
    \includegraphics[width=0.99\linewidth]{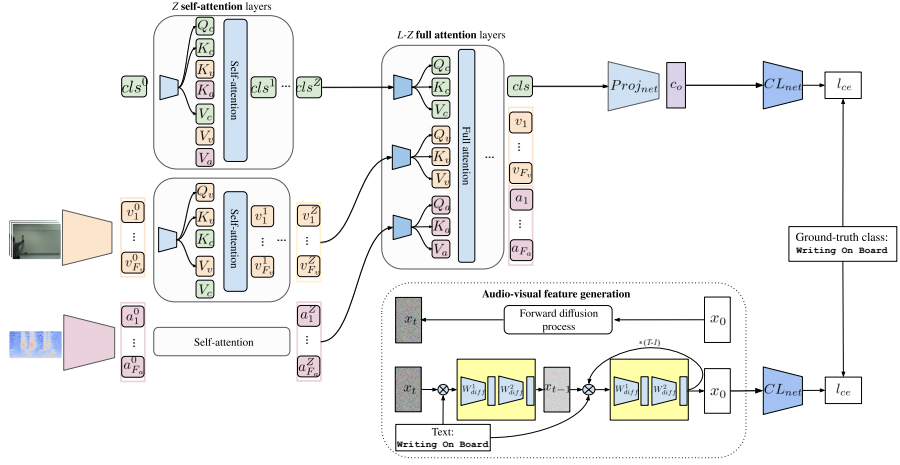}
    
    \caption{Our \ourmodel model for audio-visual (G)FSL takes audio and visual features extracted from pre-trained audio and video classification models as inputs. During training, the features from both modalities are fused into a classification token, denoted by $cls$. At the same time, our diffusion model (bottom) generates additional synthetic features for the novel classes (denoted by $x_0$). Finally, we train our classifier $CL_{net}$ (right) on fused real features $c_o$ of both novel and base classes and synthetic features of novel classes. $\otimes$ is the concatenation operator. 
    }
    
    \label{fig:architecture}
\end{figure*}

In this section, we provide details for our proposed cross-modal \ourmodel framework which employs cross-modal fusion (\Cref{sec:avfusion}) and a diffusion model to generate audio-visual features (\Cref{sec:diffusion}). 
Then, we describe the training curriculum in \Cref{sec:training_losses}.
Figure~\ref{fig:architecture} illustrates \ourmodel's full architecture.

\subsection{Audio-visual fusion with cross-modal attention}\label{sec:avfusion}

\mypara{Audio-visual fusion.} 
We project the audio ${a}_{[i]}$ and visual features ${v}_{[i]}$ to a shared embedding space. Then we use Fourier features \cite{tancik2020fourier} as temporal positional embeddings and modality embeddings respectively and obtain positional aware video $v^E_{t}$ and audio $a^E_{t}$ tokens for timestep $t$.
We prepend a classification token $cls^0 \in \mathbb{R}^{d_{dim}}$ 
to the audio and visual tokens. The output token $cls$ corresponding to $cls^0$ is the final fused audio-visual representation which is input to $Proj_{net}$. 
Our audio-visual fusion mechanism contains $L$ layers, 
which are based on multi-head attention~\cite{vaswani2017attention} $\text{Att}^l$, followed by a feed forward function $\text{FF}^l: \mathbb{R}^{d_{dim}} \xrightarrow{} \mathbb{R}^{d_{dim}}$. 
The input to the first layer is $x_{in}^1=[cls^0,a^E_1,\cdots,a^E_{T_a},v^E_1,\cdots,v^E_{T_v}]$.
The output of a layer is:
\begin{equation}
    x_{out}^l = \text{FF}^l(\text{Att}^l(x_{in}^l)+x_{in}^l) + \text{Att}^l(x_{in}^l)+x_{in}^l.
\end{equation}
In the following, we describe the first layer of the audio-visual fusion. The other layers work similarly.
Our input $x_{in}^1$ is projected to queries, keys and values with linear maps $s: \mathbb{R}^{d_{dim}} \xrightarrow{} \mathbb{R}^{d_{dim}}$ for $s \in \{q,k,v\}$. The outputs of the projection are written as zero-padded query, key and value features. For the keys we get:

\begin{align}
    \mathbf{K}_c &= [k({cls}^0), 0, \cdots, 0],\\
    \mathbf{K}_a &= [0,\cdots,0,k(a^E_1),\cdots,k(a^E_{F_a}), 0, \cdots, 0], \\
     \mathbf{K}_v &= [0,\cdots,0,k(v^E_1),\cdots,k(v^E_{F_v})].
\end{align}
The final keys are obtained as $\mathbf{K} = \mathbf{K}_{c} + \mathbf{K}_a + \mathbf{K}_v$. The queries and values are obtained in a similar way.
We define full attention as $\mathbf{A}=\mathbf{A}_{c} + \mathbf{A}_{cross} + \mathbf{A}_{self}$:
\begin{equation}\label{eq:attention}
\begin{gathered}
\mathbf{A}_{c} 
= \mathbf{Q}_{c} \, \mathbf{K}^T + \mathbf{K} \, \mathbf{Q}_{c}^T, \qquad 
\mathbf{A}_{cross} =
\mathbf{Q}_a \, \mathbf{K}_v^T + \mathbf{Q}_v \, \mathbf{K}_a^T, \\
\mathbf{A}_{self} 
= \mathbf{Q}_a \, \mathbf{K}_a^T + \mathbf{Q}_v \, \mathbf{K}_v^T.
\end{gathered}
\end{equation}
The novelty in the attention mechanism in \ourmodel is that it exploits a hybrid attention mechanism composed of two types of attention: within-modality self-attention and full-attention. The first $Z$ layers use self-attention $\mathbf{A}_{self}+\mathbf{A}_{c}$, the subsequent $L-Z$ layers leverage full attention $\mathbf{A}$.

\mypara{Audio-visual classification.} We project $cls$ to $\mathbb{R}^{d_{out}}$ by using a projection network, $c_{o}=Proj_{net}(cls)$. Then, we apply a classification layer to $c_{o}$, $\textit{logits}=CL_{net}(c_{o})$. Given the ground-truth labels $\textit{gt}$, we use a cross-entropy loss, $L_{ce}=CE(\textit{logits}, \textit{gt})$ to train the full architecture.

\subsection{Text-conditioned feature generation}\label{sec:diffusion}

\ourmodel uses a diffusion process to generate audio-visual features which is based on the Denoising Diffusion Probabilistic Models (DDPM)~\cite{ho2020denoising}. In particular, we condition the generation of features for novel classes on a conditioning signal, such as the word embedding (e.g.\ word2vec~\cite{mikolov2013efficient}) of a class name.
The diffusion framework consists of a forward process and a reverse process.

\textbf{The forward process} adds noise to the data sample $x_0$ for $T$ timesteps:
\begin{equation}
q(x_{1:T} | x_0) = \prod_{t = 1}^{T} q(x_t | x_{t-1})=\prod_{t = 1}^{T}\mathcal{N}\big(x_t; \sqrt{1-  \beta_t} x_{t-1}, \beta_t \mathbf{I}\big),
\end{equation}
where $\beta_1, \dots, \beta_T$ is the variance schedule.

As the \textbf{reverse process} $q(x_{t-1}|x_t)$ is intractable, we approximate it with a parameterised model $p_{\theta}$:
\begin{equation}
    p_{\theta}(x_{0:T})=p_{\theta}(x_T)\prod_{t=1}^T p_{\theta}(x_{t-1}|x_t)= p_{\theta}(x_T)\prod_{t=1}^T\mathcal{N}(x_{t-1}; \mu_{\theta}(x_t, t), \Sigma_{\theta}(x_t, t)).
\end{equation}
We condition the model on the timestep $t$ and the class label embedding w,

\begin{equation}\label{eq:conditioned_loss}
L_{\text{diff},w}=E_{x_0, t, w, \epsilon}[||\epsilon-\epsilon_{\theta}(\sqrt{\bar a_t}x_0+\sqrt{1-\bar a_t}\epsilon, w, t)||^2],
\end{equation}
where $\epsilon$ is the noise added at each timestep and $\epsilon_{\theta}$ is a model that predicts this noise.
The sample at timestep $t-1$ is obtained from timestep $t$ as:
\begin{equation}
p_{\theta}(x_{t-1}|x_t, w) = \mathcal{N}(x_{t-1};  \frac{1}{\sqrt{\alpha_{t}}}(x_t - \frac{\beta_t}{\sqrt{1-\bar\alpha_t}} \epsilon_{\theta}(x_t, w, t)), \sigma_{t}^2\mathcal{I}).
\end{equation}
The input to $\epsilon_{\theta}$ at timestep $t$ is obtained by concatenating $x_t,w$, and $t$.
We optimize $L_{\text{diff},w}$ to learn $p_{\theta}$.

\subsection{Training curriculum and evaluation }\label{sec:training_losses}
Each training stage (explained in \Cref{sec:benchmarks}) is split into two substages.
In the first substage, we train
the full architecture (the fusion mechanism, the diffusion model, $Proj_{net}$ and the classifier $CL_{net}$) on base classes $\mathcal{V}_{B_1}$ (or $\mathcal{V}_{B_2}$ in the second stage) by minimizing $L_{ce} +L_{\text{diff},w}$. 
The classifier $CL_{net}$ is trained only on real features for the base classes in $\mathcal{V}_{B_1}$ (or $\mathcal{V}_{B_2}$ for the second stage) in the first substage.

During the second substage, we freeze the fusion mechanism and continue to train the diffusion model, $Proj_{net}$ and $CL_{net}$ with the same training objective $L_{ce}+L_{\text{diff},w}$.
Here we consider both base and novel classes $\mathcal{V}_{B_1}$ and $\mathcal{V}_{N_1}$ classes (or $\mathcal{V}_{B_2}$ and $\mathcal{V}_{N_2}$ in the second stage), unlike in the first substage where we only used base classes. 
For each batch composed of real samples from novel classes, we generate a corresponding batch of the same size with synthetic samples using our diffusion model. $CL_{net}$ is then trained on real features from $\mathcal{V}_{B_1}$ (or $\mathcal{V}_{B_2}$ in the second stage) and on real and synthetic features for the classes in $\mathcal{V}_{N_1}$ (or $\mathcal{V}_{N_2}$ in the second stage). Freezing the audio-visual transformer ensures that its fusion mechanism does not overfit to the few samples from the novel classes. 

The diffusion model is not used for inference, and the output of the classifier $CL_{net}$ for $c_0$ provides the predicted score for each class (including the novel classes). The class with the highest score is selected as the predicted class.

\section{Experiments}
In this section, we first provide the implementation details for obtaining the presented results (\Cref{subsec:impl_details}). We then report results for our proposed \ourmodel in our benchmark study (\Cref{subsec:benchmark_performance}). 
Finally, we analyse the impact of different components of \ourmodel (\Cref{subsec:model_ablations}). 

\subsection{Implementation details}\label{subsec:impl_details}

\ourmodel uses features extracted from pre-trained audio and visual classification networks as inputs (details provided in the suppl.\ material).
\ourmodel is trained using $d_{dim}=300$ and $d_{out}=64$.
Our fusion network has $L=5, 4, 8$ transformer layers, the layer after which the attention changes is set to $Z=3, 2, 5$ on \activity, \ucf and \vgg respectively.  We train all models on a single NVIDIA 
 RTX 2080-Ti GPU. The first substage uses 30 epochs while the second one uses 20 epochs. We use the Adam optimizer~\cite{kingma2014adam}, and $\beta_1=0.9$, $\beta_2=0.999$, and weight decay of $1e^{-5}$. We use a learning rate of $7e^{-5}$ for \ucf and \activity, and $6e^{-5}$ for \vgg. For \activity and \ucf, we use a scheduler that reduces the learning rate by a factor of 0.1 when the performance has not improved for 3 epochs. We use a batch size of 32 for 
 \activity, and 64 for \ucf and \vgg.
 Each epoch consists of 300 batches. As \activity has very long videos, we randomly trim the number of features during training to 60. During evaluation, we also trim the videos to a maximum length of 300 features, and the trimmed features are centred in the middle of the video. 
 To reduce the bias towards base classes, we use calibrated stacking \cite{chao2016empirical} on the search space composed of the interval [0,1] with a step size of 0.1. This value is obtained on the validation dataset.

\begin{table*}[t]
\centering
\setlength{\tabcolsep}{2.5pt}
\renewcommand{\arraystretch}{1.2}
 \caption{\textbf{Our benchmark study for audio-visual (G)FSL}: 1,5,10-shot
 performance of our \ourmodel and compared methods on (G)FSL. The harmonic mean (HM) of the mean class accuracies for base and novel classes are reported for GFSL.   For the FSL performance, only the test subset of the novel classes is considered. Base, novel, and 20-shot performances are included in the suppl.\ material.}
 \resizebox{\linewidth}{!}{
 \begin{tabular}{l|ccccccHH|ccccccHH|ccccccHH}
 \toprule
\multirow{3}{*}{\textbf{Model} $\downarrow$} & \multicolumn{6}{c}{\vgg} & & & \multicolumn{6}{c}{\ucf} & & & \multicolumn{6}{c}{\activity} \\
 &\multicolumn{2}{c}{\textit{1-shot}} & \multicolumn{2}{c}{\textit{5-shot}} & \multicolumn{2}{c}{\textit{10-shot}} 
 &&
 &\multicolumn{2}{c}{\textit{1-shot}} & \multicolumn{2}{c}{\textit{5-shot}} & \multicolumn{2}{c}{\textit{10-shot}} 
&&
 &\multicolumn{2}{c}{\textit{1-shot}} & \multicolumn{2}{c}{\textit{5-shot}} & \multicolumn{2}{c}{\textit{10-shot}} 
 
 \\
 & HM & FSL & HM & FSL & HM & FSL & HM & FSL & HM & FSL & HM & FSL & HM & FSL& HM & FSL& HM & FSL& HM & FSL& HM & FSL & HM & FSL \\ 
\midrule
 Att. Fusion \cite{fayek2020large} &15.46&16.37&28.22&31.57&30.73&39.02&34.35&44.08&37.39&36.88&51.68&47.18&57.91&52.19&69.00&63.20&4.35&5.82&6.17&8.13&10.67&10.78&13.15&13.22\\ 
 Perceiver \cite{jaegle2021perceiver}  &17.97&18.51&29.92&33.58&33.65&40.73&35.24&43.77&44.12&33.73&48.60&40.47&55.33&47.86&59.59&52.66&17.34&12.53&25.75&21.50&29.88&26.46&31.67&32.21  \\
MBT \cite{nagrani2021attention}  &14.70&21.96&27.26&34.95&30.12&38.93&32.49&43.19&39.65&27.99&46.55&34.53&50.04&39.73&55.80&44.58&14.26&12.63&23.26&22.38&26.86&26.03&31.96&30.76\\
\TCAF \cite{mercea2022tcaf} &19.54&20.01&26.09&32.22&28.95&36.43&30.87&38.89&44.61&35.90&46.29&37.39&54.19&47.61&58.29&51.99&16.50&13.01&22.79&21.81&24.78&23.33&31.20&29.88\\
ProtoGan \cite{kumar2019protogan} &10.74&14.08&25.17&28.87&29.85&34.80&32.04&38.42&37.95&28.08&42.42&33.63&51.01&40.68&54.57&50.48&2.77&4.40&2.67&7.81&4.05&8.81&12.32&14.65 \\
SLDG \cite{bo2020few}  &16.83&17.57&20.79&25.17&24.11&29.48&24.59&33.30&39.92&28.91&36.47&28.56&34.31&26.96&53.14&43.95&13.57&10.30&22.29&19.16&27.81&25.35&31.68&32.44 \\
TSL  \cite{xian2021generalized} &18.73&22.44&19.49&29.50&21.93&31.29&22.47&32.07&44.51&35.17&51.08&42.42&60.93&55.63&60.16&52.02&9.53&10.77&10.97&12.77&10.39&12.18&11.77&15.78\\
HiP  \cite{carreira2022hierarchical} &19.27&18.64&26.82&30.67&29.25&35.13&30.89&38.46&21.79&34.88&36.44&42.23&50.69&43.29&54.06&48.07&13.80&10.31&18.10&16.25&19.37&17.06&23.13&20.67\\
Zorro  \cite{Recasens2023ZorroTM} &18.88&21.79&29.56&35.17&32.06&40.66&33.98&43.63&44.35&34.52&51.86&42.59&58.89&49.06&62.45&57.10&14.56&11.94&23.14&21.94&27.35&26.33&31.10&30.31\\
AVCA \cite{mercea2022avca} &6.29&10.29&15.98&20.50&18.08&28.27&20.75&32.64&43.61&31.24&49.19&36.70&50.53&39.17&51.39&44.93&12.83&12.22&20.09&21.65&26.02&26.76&26.91&30.76\\
\hline
  \ourmodel &\textbf{20.31}&\textbf{22.95}&\textbf{31.19}&\textbf{36.56}&\textbf{33.99}&\textbf{41.39}&\textbf{36.04}&\textbf{44.79}&\textbf{51.50}&\textbf{39.89}&\textbf{59.96}&\textbf{51.45}&\textbf{64.18}&\textbf{57.39}&\textbf{69.61}&\textbf{64.23}&\textbf{18.47}&\textbf{13.80}&\textbf{26.96}&\textbf{23.00}&\textbf{30.86}&\textbf{27.81}&\textbf{34.68}&\textbf{32.89}  \\
 \bottomrule
 \end{tabular}
}

 \label{tab:sota_table}
 
 \end{table*}

\subsection{Audio-visual GFSL performance}\label{subsec:benchmark_performance}
 
For each of the models featured in our benchmark, we report results for three different numbers of shots, i.e.\ 1-shot, 5-shot, 10-shot on all three datasets in Table \ref{tab:sota_table}. \ourmodel outperforms all the methods across all shots and datasets for few-shot learning (FSL) and generalised few-shot learning (HM). 

For 1-shot, \ourmodel achieves a HM/FSL of 20.31\%/22.95\% vs.\ HM of 19.54\% for \TCAF and FSL score of 22.44\% for TSL on \vgg. 
On 5-shot, our model obtains a HM/FSL of 31.19\%/36.56\% vs.\ 29.92\% for the Perceiver and FSL of 35.17\% for Zorro.
Furthermore, \ourmodel yields slightly better results than the Perceiver in both HM and FSL for 10 shots, with  HM/FSL of 33.99\%/41.39\% vs.\ 33.65\%/40.73\% for the Perceiver. 
Thus, combining our hybrid attention and the diffusion model is superior to systems that rely solely on powerful attention mechanisms without incorporating generative modelling (Perceiver, \TCAF) and systems that incorporate generative modelling, but that do not employ powerful attention mechanisms (TSL, ProtoGan). 

Similar trends are observed on \ucf, while on \activity, the ranking of methods changes dramatically.
Methods that perform well on \ucf and \vgg, but which do not fully use the temporal information (e.g.\ Attention Fusion, ProtoGan and TSL) perform weakly on \activity which contains videos with varying lengths, including some very long videos, making the setting more challenging. Our \ourmodel can process temporal information effectively, resulting in robust state-of-the-art results on \activity.

Interestingly, VGGSound-FSL contains the most classes among the datasets considered, resulting in a significantly lower N (suppl. material, Tab.\ 1) than FSL. This also lowers the HM (computed from B, N). On VGGSound-FSL, methods tend to be biased towards novel classes ($\text{N} \ge \text{B}$) due to calibration \cite{chao2016empirical}. In this case, $\text{HM} \le \text{N} \le \text{FSL}$. Moreover, some baselines that were also used in audio-visual zero-shot learning~\cite{mercea2022tcaf, mercea2022avca} (e.g.\ \TCAF) exhibit significant increases in performance even in the 1-shot setting. This is expected as for 1-shot learning, one training example is used from each novel class. This reduces the bias towards base classes, leading to more balanced B and N scores, and thereby better HM and FSL results.
Base, novel, and 20-shot performances are included in the suppl.\ material.

\subsection{\ourmodel model ablations}\label{subsec:model_ablations}
Here, we analyse the benefits of the main components of \ourmodel, i.e.\ our proposed audio-visual fusion mechanism, and the diffusion model for feature generation. Furthermore, we analyse the importance of using multiple modalities, and the effect of different semantic representations.

\mypara{Audio-visual fusion mechanism.} Table \ref{tab:attention_ablation_table} ablates our cross-modal fusion mechanism for generating rich audio-visual representations. As shown in \Cref{sec:avfusion}, \ourmodel uses two types of attention: $\mathbf{A}_{self}$+$\mathbf{A}_{c}$ for the first few layers and $\mathbf{A}$ for the later layers. For \textit{Alternate \ourmodel}, we alternate the two types of attention used in \ourmodel in subsequent layers. We also show our model with $\mathbf{A}_{cross}$+$\mathbf{A}_{c}$ which is the same attention used by the SOTA audio-visual GZSL framework~\cite{mercea2022tcaf}. On \activity, \ourmodel obtains a HM/FSL of 26.96\%/23.00\% vs.\ 25.58\%/22.65\% for $\mathbf{A}_{self}$+$\mathbf{A}_{c}$. The same trend is seen on \ucf. On \vgg, we outperform \textit{Alternate \ourmodel} on HM but are slightly weaker than $\mathbf{A}_{self}$+$\mathbf{A}_{c}$ in FSL. Overall, our fusion mechanism is the best across both metrics and datasets.

\begin{table}[t]
\centering
\setlength{\tabcolsep}{4pt}
\renewcommand{\arraystretch}{1.2}
 \caption{Impact of different audio-visual fusion mechanisms in the 5-shot setting.}

 \resizebox{0.9\linewidth}{!}{
 \begin{tabular}{l|cccc|cccc|ccccc}
 \toprule
\multirow{2}{*}{\textbf{Model} $\downarrow$} &  
 \multicolumn{4}{c}{\small \vgg} & 
  \multicolumn{4}{c}{\small \ucf} 
   & \multicolumn{4}{c}{\small \activity} \\
   & B & N & HM & FSL & B & N & HM & FSL & B & N & HM & FSL \\ 
   \hline
  $\mathbf{A}$ &28.56 &31.52& 29.98 &36.55 &78.95 &42.07 &54.90 &43.75 &23.10 &22.06 &22.57 &22.53 \\
  $\mathbf{A}_{cross}+\mathbf{A}_{c}$ & 28.44&32.48&30.33  &36.85 &82.89 &44.33 &57.77 &47.02 &27.02 &21.25 &23.79 &21.98 \\
  $\mathbf{A}_{self}+\mathbf{A}_{c}$ &26.68 &33.23& 29.60&\textbf{37.06}& 50.10&44.58 &47.18 &45.03 & 31.61&21.48 &25.58 &22.65 \\
  Alternate \ourmodel &27.40 &32.60& 29.78 & 36.82&80.25 &43.01 &56.00 & 45.81&31.15 &21.57 &25.49 &22.59 \\
  \hline
   \ourmodel & 30.88 & 31.50&\textbf{31.19} & 36.56 &74.11 & 50.35& \textbf{59.96} & \textbf{51.45}&35.84 &21.61 & \textbf{26.96} & \textbf{23.00} \\
 \bottomrule
 \end{tabular}
 }

 \label{tab:attention_ablation_table}
  
 \end{table}

\begin{table}[t]
\centering
\setlength{\tabcolsep}{4pt}
\renewcommand{\arraystretch}{1.2}
 \caption{Influence of using different feature generators in the 5-shot setting. 
 }
 \resizebox{0.9\linewidth}{!}{
 \begin{tabular}{l|cccc|cccc|cccc}
 \toprule
\multirow{2}{*}{\textbf{Model} $\downarrow$} & \multicolumn{4}{c}{\small \vgg} & \multicolumn{4}{c}{\small \ucf} & \multicolumn{4}{c}{\small \activity} \\
   & B & N & HM & FSL & B & N & HM & FSL & B & N & HM & FSL \\ 
   \hline
   AV-GAN & 27.80&31.75&29.64  &36.53 &83.79 &36.20 &50.56 &37.33 & 35.12&19.53 &25.10 &21.35 \\
   \hline
    \ourmodel & 30.88 & 31.50&\textbf{31.19} & \textbf{36.56} &74.11 & 50.35& \textbf{59.96} & \textbf{51.45}& 35.84 & 21.61 & \textbf{26.96} & \textbf{23.00} \\
 \bottomrule
 \end{tabular}
 }
 \label{tab:generative_ablation_table}
 
 \end{table}

\mypara{Feature generation model.}
In Table \ref{tab:generative_ablation_table}, we investigate the impact of different generative models to produce audio-visual features for the novel classes. We compare the diffusion model in \ourmodel to a GAN similar to the one used by TSL~\cite{xian2021generalized}, which optimizes a Wasserstein GAN loss~\cite{arjovsky2017wasserstein}. On \activity, we observe that \ourmodel outperforms the GAN variant, with a HM/FSL of 26.96\%/23.00\% vs.\ 25.10\%/21.35\% for the GAN. The same can be seen on \ucf and \vgg. This shows that our generative diffusion model is better suited for audio-visual GFSL than a GAN.

\mypara{Multi-modal input.}
We explore the impact of using multi-modal inputs for \ourmodel in \Cref{tab:multi_modal_ablation_table}.
For unimodal inputs, we adapt \ourmodel to only employ full attention which is identical to self-attention in this case. On \activity, using multi-modal inputs provides a significant boost in performance compared to unimodal inputs, with a HM/FSL of 26.96\%/23.00\% vs.\ 19.01\%/17.84\% when using only visual information. The same trend can be observed on \ucf. In contrast, on \vgg, using multi-modal inputs gives stronger GFSL but slightly weaker results in FSL than using the audio modality. This might be due to the focus on the audio modality in the data curation process for VGGSound. As a result, significant portions of the visual information can be unrelated to the labelled class. Overall, the use of multi-modal inputs from the audio and visual modalities significantly boosts the (G)FSL performance for \ourmodel.

However, one interesting aspect is that using both modalities leads to better \ba and \nl performances across all three datasets. For example, on \activity, \ourmodel obtains a \ba score of 35.84\% and an \nl score of 21.61\% compared to 20.80\% and 17.49\% when using only the visual modality. On \ucf, \ourmodel achieves a score of 74.11\% for \ba and 50.35\% for \nl compared to 67.13\% and 39.18\% for the visual and audio modalities respectively. Finally, on \vgg, \ourmodel achieves a \ba score of 30.88\% and an \nl score of 31.50\% compared to 28.30\% and 30.56\% for unimodal audio inputs. This shows that using multi-modal inputs decreases the bias towards either of the metrics, leading to a more robust and balanced system.

\begin{table}[t]
\centering
\setlength{\tabcolsep}{4pt}
\renewcommand{\arraystretch}{1.2}
 \caption{Influence of using multi-modal input in the 5-shot setting.
 }
 \resizebox{0.9\linewidth}{!}{
 \begin{tabular}{l|cccc|cccc|cccc}
 \toprule
\multirow{2}{*}{\textbf{Model} $\downarrow$} & \multicolumn{4}{c}{\small \vgg} & \multicolumn{4}{c}{\small  \ucf} & \multicolumn{4}{c}{\small \activity} \\
   & B & N & HM & FSL & B & N & HM & FSL & B & N & HM & FSL \\ 
   \hline
 Audio &28.30&30.56&29.39& \textbf{36.64}&55.31 &39.18 &45.87 &44.44 &13.74 &15.23 &14.45 &17.58 \\
 Visual &7.83 &8.92& 8.35 & 9.51& 67.13& 30.70& 42.14& 30.98& 20.80&17.49 &19.01 &17.84 \\
 \hline
 \ourmodel & 30.88 & 31.50&\textbf{31.19} &  36.56 &74.11 & 50.35& \textbf{59.96} & \textbf{51.45}&35.84 &21.61 & \textbf{26.96} & \textbf{23.00} \\
 \bottomrule
 \end{tabular}
 }

 \label{tab:multi_modal_ablation_table}
  
 \end{table}

\begin{table}[t]
\centering
\setlength{\tabcolsep}{4pt}
\renewcommand{\arraystretch}{1.2}
 \caption{Influence of different semantic class representations in the 5-shot setting. 
 }
 \resizebox{0.9\linewidth}{!}{
 \begin{tabular}{l|cccc|cccc|cccc}
 \toprule
 \multirow{2}{*}{\textbf{Model} $\downarrow$} & \multicolumn{4}{c}{\small \vgg} & \multicolumn{4}{c}{\small \ucf} & \multicolumn{4}{c}{\small \activity} \\
   & B & N & HM & FSL & B & N & HM & FSL & B & N & HM & FSL \\ 
   \hline
   \ourmodel $av_{prot}$ &25.74 &33.00&28.92 &35.76&83.38 &42.46 & 56.26& 44.78& 32.22&21.50 &25.79 &22.73 \\
   \hline
    \ourmodel & 30.88 & 31.50&\textbf{31.19} &  \textbf{36.56} &74.11 & 50.35& \textbf{59.96} & \textbf{51.45}&35.84 &21.61 & \textbf{26.96} & \textbf{23.00} \\
 \bottomrule
 \end{tabular}
 }

 \label{tab:semanti_information_ablation_table}
  
 \end{table}

\mypara{Semantic class representations.}
We consider using different semantic class representations in Table \ref{tab:semanti_information_ablation_table}. 
In FSL, the most common semantic descriptor is word2vec~\cite{mikolov2013efficient} which is used to condition the audio-visual feature generation in \ourmodel.
However, related works (e.g.\ ProtoGan~\cite{kumar2019protogan}), use prototypes which average the visual features of all the training videos in a class to obtain the semantic representation of that class.
In the multi-modal setting, we can concatenate the audio and visual prototypes to obtain multi-modal prototypes $av_{prot}$ which is used as a conditioning signal for our diffusion model. On \activity, using word2vec embeddings leads to better results than using the audio-visual prototypes $av_{prot}$, with a HM/FSL of 26.96\%/23.00\% vs.\ 25.79\%/22.73\% for $av_{prot}$. The same can be seen on \ucf and \vgg, demonstrating that the word2vec embeddings provide a more effective conditioning signal.

\section{Conclusion}
In this work, we propose an audio-visual (generalised) few-shot learning benchmark for video classification. Our benchmark includes training and evaluation protocols on three datasets, namely \vgg, \ucf and \activity, and baseline performances for ten state-of-the-art methods adapted from different fields.
Moreover, we propose \ourmodel which fuses multi-modal information with a hybrid attention mechanism and uses a text-conditioned diffusion model to generate features for novel classes. \ourmodel outperforms all related methods on the new benchmark. Finally, we provided extensive model ablations to show the benefits of our model's components. We hope that our benchmark will enable significant progress for audio-visual generalised few-shot learning.

\vspace{2em}
\mypara{Acknowledgements:} This work was supported by BMBF FKZ: 01IS18039A, DFG: SFB 1233 TP 17 - project number 276693517, by the ERC (853489 - DEXIM), and by EXC number 2064/1 – project number 390727645. The authors thank the International Max Planck Research School for Intelligent Systems (IMPRS-IS) for supporting O.-B. Mercea and T. Hummel.

\bibliographystyle{splncs04}
\bibliography{102-main.bib}

\makeatletter

\makeatother

\renewcommand\UrlFont{\color{blue}\rmfamily}

%
%
%

\newif\ifreview
\reviewfalse

\ifreview
	\usepackage{lineno}
	\renewcommand\thelinenumber{\color[rgb]{0.2,0.5,0.8}\normalfont\sffamily\scriptsize\arabic{linenumber}\color[rgb]{0,0,0}}
	\renewcommand\makeLineNumber {\hss\thelinenumber\ \hspace{6mm} \rlap{\hskip\textwidth\ \hspace{6.5mm}\thelinenumber}} 
	\linenumbers
\fi


\def\SubNumber{102}

\def\GCPRTrack{Fast Track, Supplementary material}

\title{Supplementary: Text-to-feature diffusion for audio-visual few-shot learning}

\ifreview
	\titlerunning{GCPR 2023 Submission \SubNumber{}. CONFIDENTIAL REVIEW COPY.}
	\authorrunning{GCPR 2023 Submission \SubNumber{}. CONFIDENTIAL REVIEW COPY.}
	\author{GCPR 2023 - \GCPRTrack{}}
	\institute{Paper ID \SubNumber}
\else

	\author{Otniel-Bogdan Mercea\inst{1}\orcidlink{0000-0002-3586-1703}\index{Mercea, Otniel-Bogdan} \and
	Thomas Hummel\inst{1}\orcidlink{0000-0003-3201-360X} \and
	A. Sophia Koepke\inst{1}\orcidlink{0000-0002-5807-0576}\index{Koepke, A. Sophia} \and Zeynep Akata\inst{1,2}\orcidlink{0000-0002-1432-7747}}
	
	\authorrunning{O.-B. Mercea et al.}
	
	\institute{University of T{\"u}bingen \and MPI for Intelligent Systems \\
	\email{\{otniel-bogdan.mercea, thomas.hummel, a-sophia.koepke,zeynep.akata\}@uni-tuebingen.de}}
\fi

\maketitle              
In Section \ref{section_extracting_features}, we describe the procedure used to extract the audio and visual features that are used as inputs to our \ourmodel framework. In Section \ref{additional_experimental_results}, we provide additional experimental results for (G)FSL with 20 shots, along with reporting the GFSL performance on base and novel classes across all shots and datasets. Finally, we provide additional ablations on the hybrid attention and diffusion model.

\section{Feature extraction} \label{section_extracting_features}

We train \ourmodel on already pre-extracted temporal features for the audio and visual modalities. We used C3D \cite{tran2015learning} which was pre-trained on Sports1M ~\cite{karpathy2014sports} 
and VGGish \cite{hershey2017cnn} pre-trained on YouTube-8M~\cite{abu2016youtube} to extract audio and visual features respectively.
Each audio feature is represented by a 128-dimensional vector corresponding to one second of audio data.
To extract the visual features, we first resampled the videos to 25fps and then extracted a 4096-dimensional vector for 16 consecutive video frames.   

\section{Additional experimental results} \label{additional_experimental_results}
We present (G)FSL results for 20 shots on the \ucf, \vgg and \activity datasets in Section~\ref{20_shots_results}. In Section \ref{b_and_n}, we discuss the 1-,5-,10- and 20-shot (G)FSL performance on base and novel  classes across all three datasets (which complements Section 5.2 of the main paper). Finally, Section \ref{additional_abl} shows additional ablations on the hybrid attention and diffusion model.

\subsection{(G)FSL in the 20-shot setting}\label{20_shots_results}

In Table \ref{tab:final_results_base_novel} (bottom), we provide additional (G)FSL results for the 20-shot setting with \ourmodel and related methods.
Similar to our observations in the main paper with 1, 5, and 10 shots, \ourmodel achieves state-of-the-art performance for 20 shots, outperforming all related methods in the FSL and GFSL (HM) settings. 

Similar to the conclusions for \activity in the main paper, it can be observed that the ranking of baselines changes dramatically on \activity, while \ourmodel still remains the best, showing that our model is also more robust on 20 shots.

The HM and FSL performances on 20 shots for \ourmodel and for the related methods are higher compared to the lower shots. The increase in performance for \ourmodel from 10 to 20 shots is similar to the one from 5 to 10 shots. However, the most significant boost in performance happens between the 1-shot and 5-shot settings, showing that the gain in performance decreases as more training samples for novel classes are added. Similar trends can also be observed for the related methods.

\subsection{Performance on base and novel classes}\label{b_and_n}

In the main paper, we only presented the GFSL results in terms of the harmonic mean of the performance on the \ba (base) and \nl (novel) classes (Table 2 in the main paper). The harmonic mean is crucial as it evaluates how robust a system is, and it also provides higher scores to systems which are very balanced and which are less biased towards either \ba or \nl. 
In this section, we are going to analyse the performance of the components that are used to calculate the HM, namely the \ba and \nl performance, to have a better idea of the model's strengths and weaknesses. It can be seen in Table \ref{tab:final_results_base_novel} that in the majority of cases, \ourmodel obtains state-of-the-art performance on \ba and \nl, but there are still some exceptions, as presented below.

In the 1-shot setting, it can be observed that MBT outperforms \ourmodel on \nl in \vgg and \ba in \ucf, with scores of 21.34\% and 79.89\% compared to 21.25\% and 77.94\% for \ourmodel. However, MBT is very biased towards one of the metrics. On \vgg, the bias is towards \nl, and MBT obtains a very low score on \ba, only 11.21\%, compared to 19.44\% for \ourmodel. The same applies to \ucf, where MBT is very biased towards \ba. 
For \ba on \vgg, \ourmodel obtains a performance of 19.44\% compared to 28.55\% SLDG. While \ourmodel scores similarly on both metrics in \vgg, SLDG obtains a \ba score which is more than twice that of \nl, showing how unbalanced and biased SLDG is. An interesting observation that can be made in the 1-shot setting is that on \vgg, \ourmodel is not able to attain state-of-the-art performance in \ba or \nl, but it still performs overall much better than the systems that outperform \ourmodel in these two metrics.

In the 5-shot setting, \ourmodel is outperformed on \ba in both \vgg and \ucf by the Perceiver, with scores of 31.46\% and 83.56\% compared to 30.88\% and 74.11\% for \ourmodel. Moreover, on \vgg, \ourmodel is also outperformed on \nl by MBT with scores of 31.79\% vs 31.50\% for \ourmodel. However, both MBT and Perceiver have a bigger bias towards one of the metrics, leading to a lower HM on \vgg. On \ucf, it can be clearly observed that Perceiver is biased towards \ba, obtaining a score which is more than twice that of \nl. For \ourmodel this is not the case, as scores for both \ba and \nl are much more balanced.

The same observations can be made in the 10- and 20-shot settings where sometimes \ourmodel is outperformed in one of the \ba or \nl, but still achieves a higher HM overall.
While most of the baselines that outperform \ourmodel in one of the metrics are usually very biased towards that metric, this is not always the case. For example, in the 20-shot setting on \ucf, Att.\ Fusion slightly outperforms \ourmodel on \nl with a score of 61.02\% compared to 59.94\% for \ourmodel. However, on \ba, \ourmodel significantly outperforms Att.\ Fusion with a score of 86.51\% compared to 79.39\% for Att.\ Fusion. While in this case Att.\ Fusion is very well balanced, it is still worse overall than \ourmodel, as it only slightly outperforms \ourmodel in \nl but it is significantly outperformed in \ba.

Interestingly, for different methods, the \nl score is sometimes higher than \ba. This is likely due to the use of calibrated stacking \cite{chao2016empirical}. Similar behaviour has been observed by several other works, such as \cite{mercea2022avca,mercea2022tcaf,min2020domain} 

Overall, \ourmodel is not necessarily the best in both \ba and \nl every single time. However, across all shots and datasets, \ourmodel achieves state-of-the-art GFSL performance in terms of the HM. This shows that \ourmodel is the most balanced and robust among all the methods, as it can consistently score very high on both \ba and \nl.

\begin{table}[t]
\centering
\setlength{\tabcolsep}{4pt}
\renewcommand{\arraystretch}{1.2}
\caption{\textbf{Novel (N) and base (B) performance for audio-visual (G)FSL}: 1-shot, 5-shot, 10-shot, and 20-shot
 performance of \ourmodel and compared methods on the \vgg, \ucf and \activity datasets. The harmonic mean (HM) of the mean class accuracies for base and novel classes are reported for GFSL.  The FSL performance considers only the test subset of novel classes.}
 \resizebox{\linewidth}{!}{
 \begin{tabular}{l|cccc|cccc|cccc}
 \toprule
\multirow{2}{*}{\textbf{1-shot}}  & \multicolumn{4}{c}{\vgg} & \multicolumn{4}{c}{\ucf} & \multicolumn{4}{c}{\activity} \\
   & B & N & HM & FSL & B & N & HM & FSL & B & N & HM & FSL \\ 
   \hline
 Att. F. \cite{fayek2020large} & 15.16&15.77&15.46&16.37&38.91 & 35.98 &37.39 &36.88 & 3.48 & 5.78&4.35 &5.82   \\
 Perc. \cite{jaegle2021perceiver} &18.46 & 17.51& 17.97&18.51 & 74.57&31.33 & 44.12 & 33.73 &30.32&12.14 & 17.34 & 12.53  \\ 
  MBT \cite{nagrani2021attention} &11.21&21.34&14.70&21.96&79.89&26.37&39.65&27.99&17.07&12.24&14.26&12.63 \\
 \TCAF \cite{mercea2022tcaf} & 20.93& 18.34& 19.54& 20.01& 66.18 & 33.64& 44.61& 35.90& 23.85&12.62 &16.50 &13.01  \\
Proto \cite{kumar2019protogan} &8.85&13.65&10.74&14.08&60.12&27.72&37.95&28.08&2.02&4.40&2.77&4.40 \\
SLDG \cite{bo2020few} &28.55&11.94&16.83&17.57&73.15&27.45&39.92&28.91&23.22&9.58&13.57&10.30 \\
 TSL~\cite{xian2021generalized} & 17.09&20.72& 18.73&22.44 & 68.18&33.04 &44.51&35.17 &8.96&10.18&9.53&10.77   \\
 HiP~\cite{carreira2022hierarchical} & 23.39 & 16.39 & 19.27 & 18.64 & 16.20  & 33.26 & 21.79 & 34.88 & 25.02 & 9.53 & 13.80 & 10.31 \\
 Zorro~\cite{Recasens2023ZorroTM} & 17.49 & 20.51 & 18.88 & 21.79 & 67.85 & 32.94 & 44.35 & 34.52 & 19.67 & 11.55 & 14.56 &11.94 \\
AVCA~\cite{mercea2022avca} &4.53 & 10.28& 6.29& 10.29& 82.86 & 29.59& 43.61 &31.24&14.15 & 11.73 & 12.83 & 12.22 \\
\hline
  \ourmodel & 19.44 &21.26 & \textbf{20.31} & \textbf{22.95} &77.94 &38.46 & \textbf{51.50} & \textbf{39.89}  & 32.77 & 12.86 &\textbf{18.47} &\textbf{13.80} \\
 \bottomrule
 \toprule
 \multirow{2}{*}{\textbf{5-shot}} & \multicolumn{4}{c}{\vgg} & \multicolumn{4}{c}{\ucf} & \multicolumn{4}{c}{\activity} \\
   & B & N & HM & FSL & B & N & HM & FSL & B & N & HM & FSL \\ 
   \hline
 Att. F. \cite{fayek2020large} & 28.64 & 27.82 & 28.22&31.57 & 63.27& 43.69&51.68 & 47.18 &5.00&8.05 &6.17 &8.13   \\
 Perc. \cite{jaegle2021perceiver} &31.46 & 28.52 & 29.92 &33.58 &83.56 &34.27 &48.60 &40.47 & 35.66 & 20.15 &25.75 &21.50  \\ 
 MBT \cite{nagrani2021attention} &23.86&31.79&27.26&34.95&80.61&32.72&46.55&34.53&25.36&21.48&23.26&22.38 \\
 \TCAF \cite{mercea2022tcaf} & 24.34 &28.11 &26.09 &32.22 & 73.76 & 33.73 & 46.29& 37.39& 24.45 & 21.35 & 22.79 & 21.81 \\
 Proto \cite{kumar2019protogan} &25.27&25.08&25.17&28.87 &63.69&31.79&42.42&33.63& 1.61&7.81&2.67&7.81 \\
 SLDG \cite{bo2020few} &29.74&15.98&20.79&25.17&65.44&25.28&36.47&28.56&29.40&17.95&22.29&19.16 \\
 TSL \cite{xian2021generalized} &15.02&27.75&19.49&29.50&68.80&40.62&51.08&42.42 &9.93&12.27&10.97&12.77 \\
  HiP~\cite{carreira2022hierarchical} & 30.01 & 24.18 & 26.82 & 30.67 & 33.65 & 39.74 & 36.44 & 42.23 & 21.98 & 15.39 & 18.10 & 16.25 \\
  
Zorro~\cite{Recasens2023ZorroTM} & 29.06 & 30.07 & 29.56 & 35.17 & 69.13& 41.49 & 51.86 & 42.59 & 25.72 &  21.03&  23.14&21.94 \\
AVCA~\cite{mercea2022avca} & 13.24& 20.15&15.98 &20.50 & 84.80& 34.64& 49.19 & 36.70& 19.18& 21.09 &20.09  & 21.65 \\
\hline
  \ourmodel & 30.88 & 31.50&\textbf{31.19} &  \textbf{36.56} &74.11 & 50.35& \textbf{59.96} & \textbf{51.45}&35.84 &21.61 & \textbf{26.96} & \textbf{23.00} \\
 \bottomrule
 \toprule
\multirow{2}{*}{\textbf{10-shot}} & \multicolumn{4}{c}{\vgg} & \multicolumn{4}{c}{\ucf} & \multicolumn{4}{c}{\activity} \\
   & B & N & HM & FSL & B & N & HM & FSL & B & N & HM & FSL \\ 
   \hline
 Att. F. \cite{fayek2020large} &26.87 & 35.89& 30.73& 39.02& 73.53&47.77 & 57.91& 52.19& 12.58& 9.27& 10.67 &10.78   \\
 Perc. \cite{jaegle2021perceiver} & 32.64& 34.73 & 33.65& 40.73& 71.88& 44.97 & 55.33 &47.86 & 37.06&25.03 & 29.88 & 26.46  \\ 
 MBT \cite{nagrani2021attention} &26.76&34.43&30.12&38.93&84.07&35.62&50.04&39.73&29.06&24.98&26.86&26.03 \\
\TCAF \cite{mercea2022tcaf} & 26.62&31.73 &28.95 &36.43 & 84.28 & 39.93 & 54.19& 47.61& 27.86&22.32 &24.78 &23.33 \\
Proto \cite{kumar2019protogan} & 30.48&29.26& 29.85& 34.80& 70.28& 40.03&51.01&40.68&2.63&8.81&4.05&8.81 \\
SLDG \cite{bo2020few} & 28.32 &20.99 &24.11 &29.48 &49.35&26.29&34.31&26.96&34.69&23.20&27.81&25.35 \\
 TSL \cite{xian2021generalized} & 17.96&28.15& 21.93&31.29&74.31&51.63&60.93&55.63 &9.31&11.76&10.39 &12.18 \\
 HiP~\cite{carreira2022hierarchical} & 28.43 & 30.12 & 29.25 & 35.13 & 75.54 & 38.14 & 50.69 & 43.29 & 24.32 & 16.10 & 19.37 & 17.06 \\
Zorro~\cite{Recasens2023ZorroTM} & 28.48 & 36.68 & 32.06 & 40.66 & 82.88 & 45.67 & 58.89 & 49.06 & 30.11 & 25.05 & 27.35 &26.33 \\
AVCA~\cite{mercea2022avca} &13.39 &27.83 &18.08 & 28.27& 71.96& 38.93& 50.53 & 39.17& 26.36& 25.68 & 26.02 &26.76  \\
\hline
  \ourmodel & 32.15 &36.05& \textbf{33.99} &\textbf{41.39}& 84.62&51.69& \textbf{64.18}& \textbf{57.39}&37.91 &26.02& \textbf{30.86} & \textbf{27.81}  \\
 \bottomrule
 \toprule

\multirow{2}{*}{\textbf{20-shot}} & \multicolumn{4}{c}{\vgg} & \multicolumn{4}{c}{\ucf} & \multicolumn{4}{c}{\activity} \\
   & B & N & HM & FSL & B & N & HM & FSL & B & N & HM & FSL \\ 
   \hline
 Att. F. \cite{fayek2020large} & 31.43& 37.88& 34.35& 44.08 & 79.39& 61.02& 69.00& 63.20 & 15.51&11.41 & 13.15 & 13.22\\ 
 Perc. \cite{jaegle2021perceiver} & 33.11& 37.66& 35.24 & 43.77& 77.81& 48.29& 59.59&52.66 &32.30 & 31.06 & 31.67&32.21  \\
MBT \cite{nagrani2021attention} &28.41&37.95&32.49&43.19&81.73&42.35&55.80&44.58&36.21&28.60&31.96&30.76 \\
\TCAF \cite{mercea2022tcaf} &32.48 & 29.41&30.87 & 38.89& 75.71 & 47.38& 58.29& 51.99& 35.87&27.61 & 31.20&29.88 \\
 Proto \cite{kumar2019protogan} & 31.44&32.66&32.04 & 38.42& 61.07& 49.32&54.57& 50.48&25.05&8.17&12.32&14.65 \\
 SLDG \cite{bo2020few} &33.20&19.53&24.59&33.30&81.08&39.52&53.14&43.95&32.60&30.80&31.68&32.44 \\
 TSL  \cite{xian2021generalized} & 18.21&29.32& 22.47& 32.07& 76.82& 49.44&60.16&52.02 &9.68&15.01&11.77 &15.78  \\
 HiP~\cite{carreira2022hierarchical} & 32.03 & 29.83 & 30.89 & 38.46 & 71.59 & 43.43 & 54.06 & 48.07 & 33.78 & 17.59 & 23.13 & 20.67 \\
Zorro~\cite{Recasens2023ZorroTM} & 29.84 & 39.46& 33.98 & 43.63 & 87.82 & 48.46 & 62.45 & 57.10 & 34.15 & 28.55 & 31.10 &30.31 \\

AVCA~\cite{mercea2022avca}&15.30 &32.20 &20.75 &32.64 & 60.00& 44.93& 51.39&44.93 &24.47 & 29.88  & 26.91 &30.76  \\
 
\hline
  \ourmodel &33.17 &39.46& \textbf{36.04} & \textbf{44.79} &86.51 &59.94& \textbf{70.82}& \textbf{65.73} & 39.25&31.06&\textbf{34.68}&\textbf{32.89}  \\

 \bottomrule
 \end{tabular}
 }
 \label{tab:final_results_base_novel}
\end{table}

\subsection{Ablation on hybrid attention and diffusion.} \label{additional_abl}
\begin{figure*}[t]
    \centering
    \includegraphics[width=0.60\linewidth]{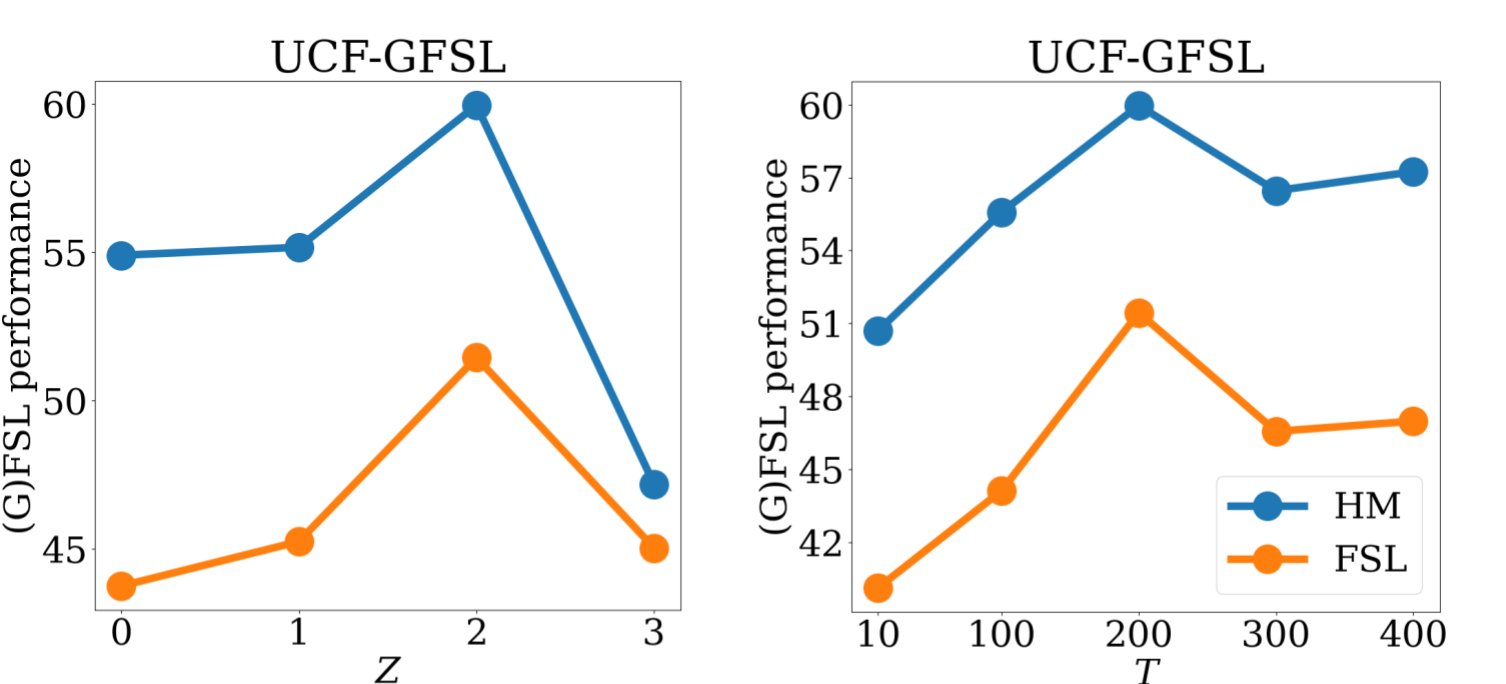}
    \caption{(G)FSL 
  performance (5-shot) for different numbers of self- ($Z$) and full attention layers (\textit{left}), and different amounts of noise addition time steps $T$ on \ucf (\textit{right}).
    }
    \label{ablation_diffusion_timestep}
\end{figure*}

In Fig \ref{ablation_diffusion_timestep} (left), we analyse the impact of the number of self-attention layers $Z$ and full-attention layers used. For values of $Z<2$ the performance increases consistently and reaches a peak performance at $Z=2$ for both metrics on \ucf. It appears that changing the attention in late layers of the network is beneficial.
Finally, we ablate over the timesteps $T$ for adding noise to the original feature in the diffusion model in Fig \ref{ablation_diffusion_timestep} (right). The (G)FSL performance maximizes for $T=200$ on \ucf which corresponds to the number of timesteps used in \ourmodel. 

\end{document}